\documentclass[letterpaper, 10 pt, conference]{ieeeconf}  

\IEEEoverridecommandlockouts                              

\usepackage{graphicx}
\usepackage{times} 
\usepackage{amsmath} 
\usepackage{amssymb}  
\usepackage{multirow}
\usepackage{multicol}
\usepackage{colortbl}
\usepackage{booktabs}
\usepackage{caption}
\usepackage{subcaption}
\usepackage{tabularx}
\usepackage{float}
\usepackage{amsfonts}
\usepackage{verbatim}
\usepackage{pifont}
\usepackage[dvipsnames]{xcolor}
\usepackage{tablefootnote}
\usepackage{scrextend}
\usepackage{hyperref}
\usepackage{cite}

\let\labelindent\relax
\usepackage[inline]{enumitem}

\usepackage{dsfont}

\definecolor{mygreen}{rgb}{0.0, 0.5, 0.0}
\definecolor{myred}{rgb}{0.82, 0.1, 0.26}

\newcommand{\cmark}{\ding{51}}%
\newcommand{\xmark}{\ding{55}}%

\usepackage{color}
\usepackage{xcolor}
\usepackage[textsize=small]{todonotes}
\definecolor{Review}{rgb}{1.0,0.9,0.9}
\definecolor{ReviewAddressed}{rgb}{0.9,0.9,1.0}
\newcommand{\commentthis}[3]{\todo[inline, color={#1}]{  {\textbf{#2}: #3}  }}
\newcommand{\review}[2] { \commentthis{Review}{R#1}{"#2"}}
\newcommand{\reviewaddr}[2] { \commentthis{ReviewAddressed}{R#1}{"#2"}}
\renewcommand{\review}[2] {}
\renewcommand{\reviewaddr}[2] {}

\usepackage[acronym]{glossaries}
\newacronym{cnn}{CNN}{Convolutional Neural Network}
\newacronym{miou}{mIoU}{mean Intersection over Union}
\newacronym{iou}{IoU}{Intersection over Union}
\newacronym{fid}{FID}{Frech\'et Inception Distance}
\newacronym{gan}{GAN}{Generative Adversarial Network}
\newacronym{cg}{CG}{Computer Graphics}
\newacronym{sfm}{SfM}{Structure from Motion}
\newacronym{rmse}{RMSE}{Root Mean Squared Error}
\newacronym{mse}{MSE}{Mean Squared Error}
\newacronym{gt}{GT}{ground-truth}
\newacronym{sota}{SOTA}{state-of-the-art}

\usepackage{cleveref}
\crefname{table}{Tab.}{Tabs.}
\crefname{figure}{Fig.}{Figs.}
\crefname{section}{Sec.}{Secs.}
\crefname{equation}{Eq.}{Eqs.}
\crefname{algorithm}{Alg.}{Algs.}

\usepackage{copyright}

\newcommand*{\figuretitle}[1]{%
    {\centering
    \textbf{#1}
    \par\medskip}
}

\renewcommand{\baselinestretch}{0.95}

\title{\LARGE \bf Depth-SIMS: Semi-Parametric Image and Depth Synthesis}

\author{Valentina Mușat$^{1}$, Daniele De Martini$^{*}$, Matthew Gadd$^{*}$, Paul Newman
\\
Mobile Robotics Group (MRG), University of Oxford\\\texttt{\{valentina,daniele,mattgadd,pnewman\}@robots.ox.ac.uk}
\thanks{$^{1}$Corresponding author.
        $^{*}$Equal contribution.
        Thanks to the Assuring Autonomy International Programme, a partnership between Lloyd’s Register Foundation and the University of York, as well as EPSRC Programme Grant ``From Sensing to Collaboration'' (EP/V000748/1).
        }%
}

\begin{document}

\maketitle
\thispagestyle{empty}
\pagestyle{empty}

\copyrightnotice

\begin{abstract}

In this paper we present a compositing image synthesis method that generates RGB canvases with well aligned segmentation maps and sparse depth maps, coupled with an in-painting network that transforms the RGB canvases into high quality RGB images and the sparse depth maps into pixel-wise dense depth maps. We benchmark our method in terms of structural alignment and image quality, showing an increase in mIoU over SOTA by 3.7 percentage points and a highly competitive FID. Furthermore, we analyse the quality of the generated data as training data for semantic segmentation and depth completion, and show that our approach is more suited for this purpose than other methods.

\end{abstract}

\glsresetall

\section{Introduction}

Vision for autonomous driving has come a long way due to the availability of high-quality datasets with manual \gls{gt} annotations \cite{yu2020bdd100k,cityscapes,a2d2}.
Still, providing \gls{gt} annotation for real datasets is cumbersome, expensive and slow, especially at pixel-level.
To overcome this bottleneck, several methods have emerged, enabled by the rapid development of deep-learning, two of which have received significant attention - domain adaptation and image synthesis. 


While domain adaptation seeks to adapt data from a source domain  to resemble the characteristics of a target domain (in particular day-to-night and sim-to-real appearance style transfer), image synthesis aims to generate data from abstract representations instead, such as semantic segmentation or latent embeddings. 
In the context of image synthesis, end-to-end parametric approaches involving \glspl{gan} have been particularly successful at reproducing the spatial structure of segmentation layouts, thus scoring high on metrics such as \gls{miou}.  On the other hand, semi-parametric approaches that combine both compositing-based and learned methods \cite{qi2018}, achieve \gls{sota} results in terms of perceptual image quality, as measured by the \gls{fid}.


\reviewaddr{1}{ "examining whether a hight score ... in mIoU leads to better results when using generated data to train downstream task". Does this mean that results on downstream task lead to better performance on downstream task? Please, clarify }

\reviewaddr{1}{ "results in .. structural alignment of the synthesised image". How does this quantified and measured? }

In this paper, we look at two strong and well-known examples of the above methods in the context of training data synthesis.  We investigate the effect of perceptual image quality and structural alignment (measured via \gls{fid} and \gls{miou} respectively) of the synthesised images on the performance of semantic segmentation and depth completion.
In doing so, we propose a set of improvements to a compositing framework derived from SIMS \cite{qi2018}, yielding \gls{sota} results in terms structural alignment of the synthesised images with their \gls{gt}, highly-competitive results in terms of perceptual quality, \emph{as well as} surpassing previous methods in terms of performance on downstream tasks trained with the generated data. Moreover, since our synthesised images are well aligned with their segmentation maps, we extend the model to also synthesise dense depth from sparse depth information, such that datasets with LiDAR information can be leveraged.

\begin{figure}[t]
\centering
     \begin{subfigure}[b]{0.23\textwidth}
         \centering
         \includegraphics[width=\textwidth]{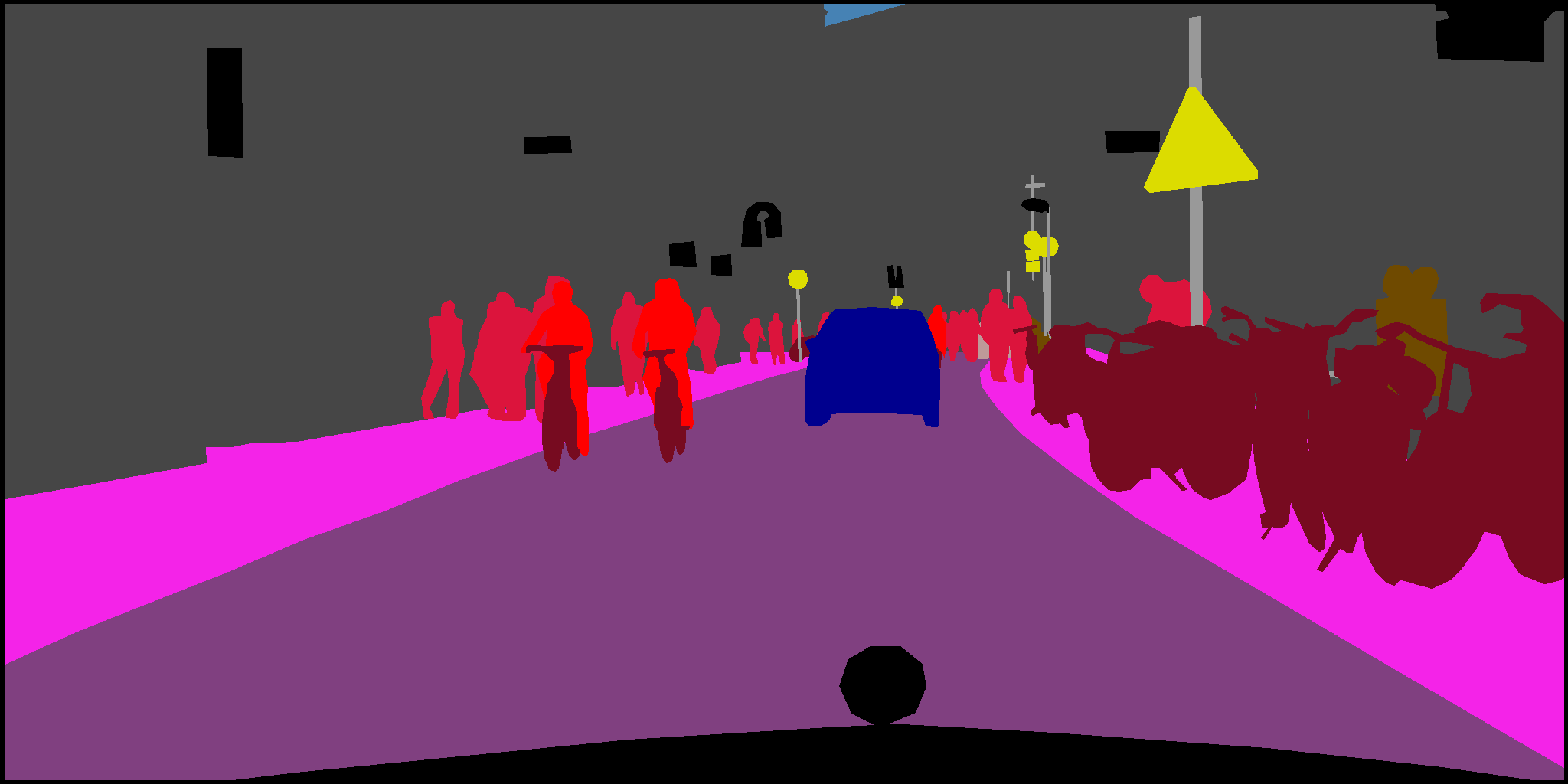}
         \caption{Input segmentation}
     \end{subfigure}
     \begin{subfigure}[b]{0.23\textwidth}
         \centering
         \includegraphics[width=\textwidth]{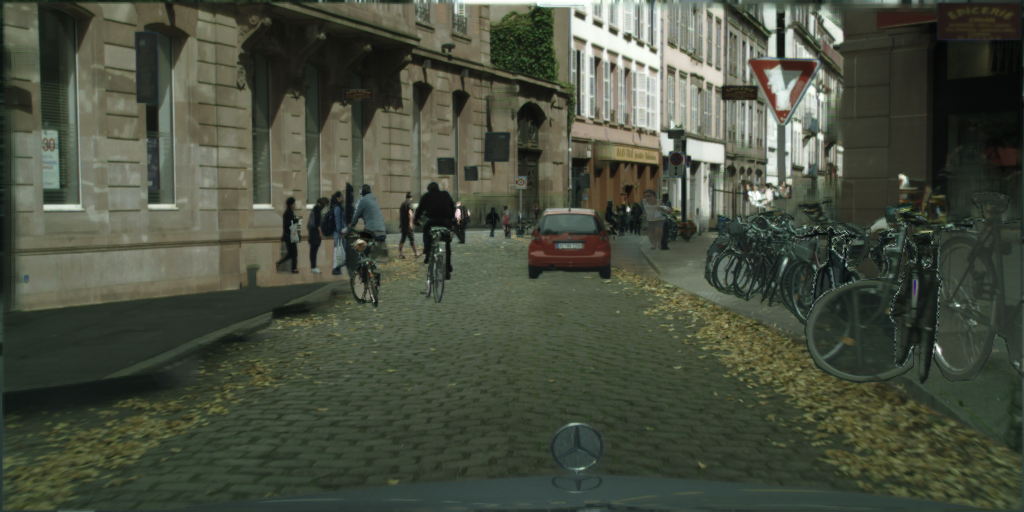}
         \caption{In-painted RGB canvas}
     \end{subfigure}
     
     \begin{subfigure}[b]{0.23\textwidth}
         \centering
         \includegraphics[width=\textwidth]{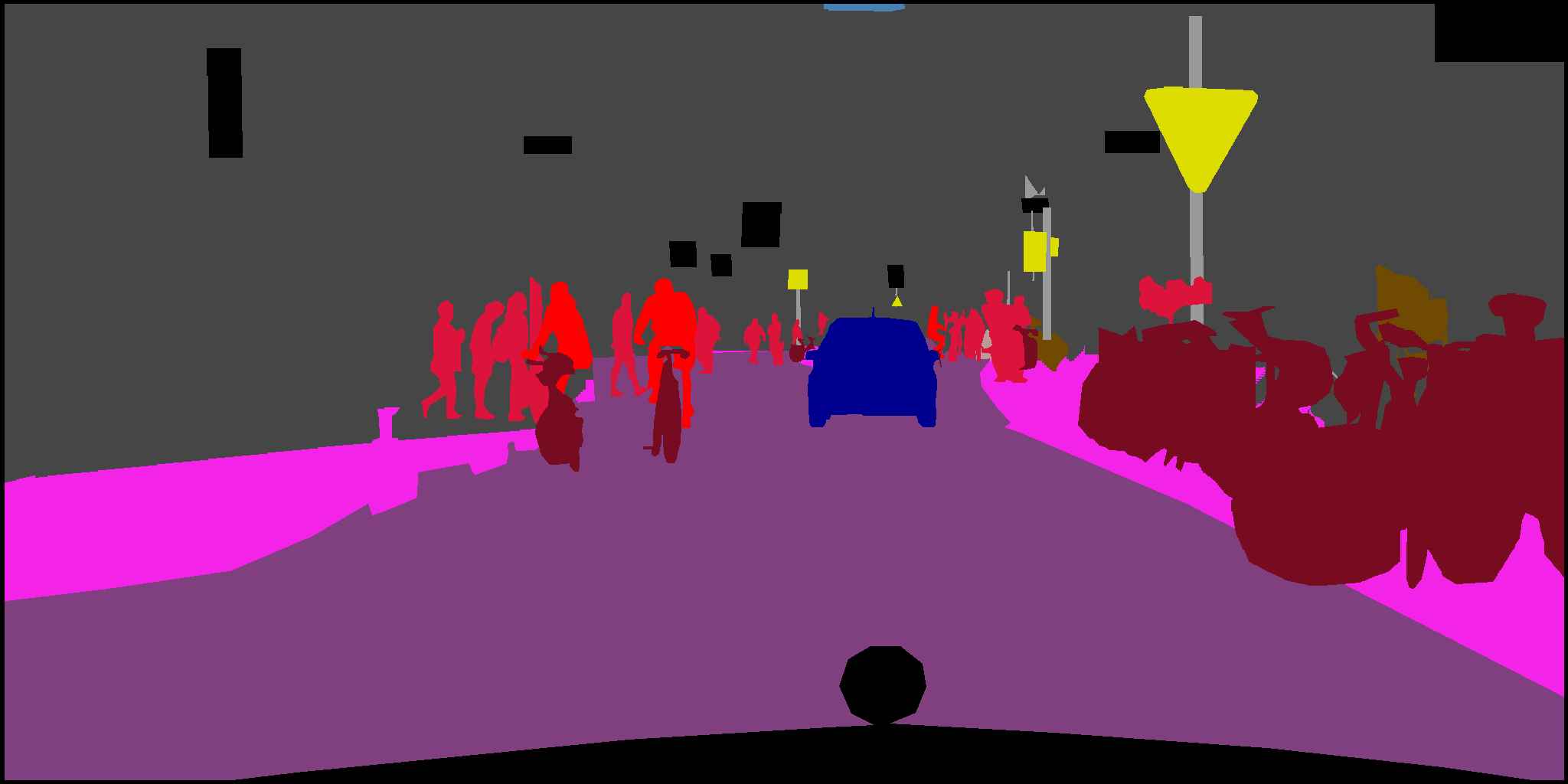}
         \caption{Output segmentation}
     \end{subfigure}
     \begin{subfigure}[b]{0.23\textwidth}
         \centering
         \includegraphics[width=\textwidth]{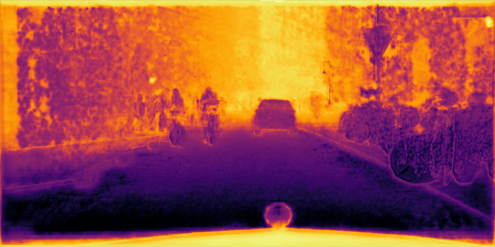}
         \caption{Synthesised depth}
     \end{subfigure}
    \caption{\small Depth-SIMS produces in-painted RGB canvasses \emph{alongside} synthesised dense depth maps.}
    \label{fig:sims_depth_example}
    \vspace{-0.6cm}
\end{figure}



\reviewaddr{2}{ In Fig. 2 and Fig. 5, some text is barely readable in the printed version }

\reviewaddr{3}{I do feel as though Fig. 2 could be polished and made clearer, particularly because there are lots of processes in the proposed approach. Maybe make use of the horizontal space to not consume more pages. The text could be larger, the rounding radii could be smaller. The arrows could be arranged to have fewer intersections The same is true of Fig.5  }

\begin{figure*}[htbp]
    \centering
\vspace{55pt}
  \includegraphics[scale=0.5, trim=20.0cm 8.0cm 15.0cm 12.5cm]{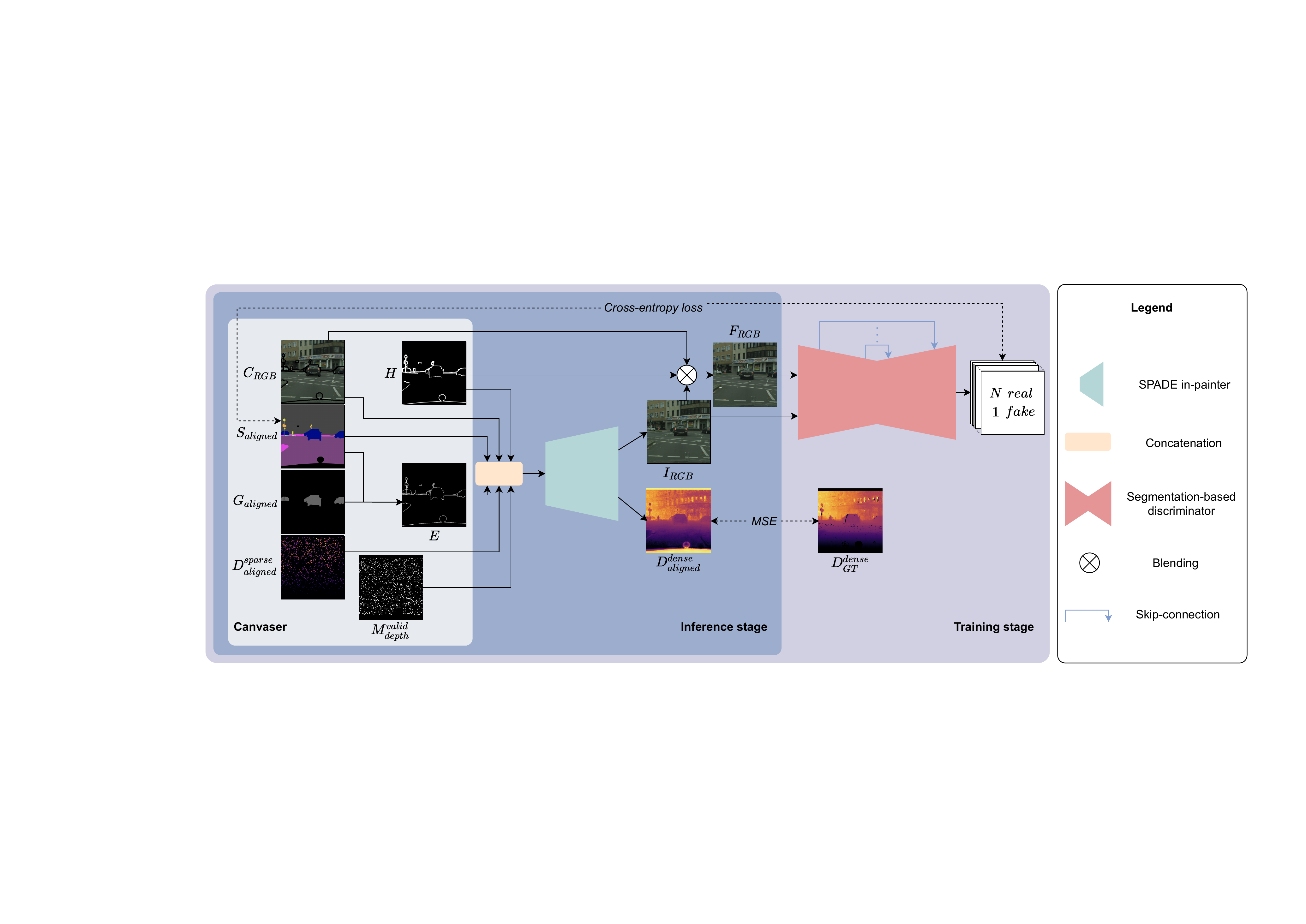}
  \caption{\small
GAN-based in-painting of RGB canvas holes and densification of sparse canvas depth maps; the blue area indicates the processes active at inference time, while the purple area shows the components used for the training procedure.
We use a generator based on SPADE~\cite{park2019} to produce in-painted RGB images, $I_{RGB}$, \emph{as well as} a dense depth map by giving it access to an RGB canvas, hole-and-boundary map, the semantic canvas, the edges of blob instances and finally sparse depth maps and masks -- which are shown and described further in~\cref{sec:canvasing,fig:canvasing}.
 }
  \label{fig:architecture}
  \vspace{-20pt}
\end{figure*}

\reviewaddr{1}{Fig.2 legend unclear (box colors and outlines)}

\review{1}{Actual contribution on top of the SIMS concept except adding depth GT is not clear}

\review{2}{ In SIMS, an ordering network is employed to determine in which order blobs are put on the canvas. The authors replaced this by a class-based order. It remains unclear why they do so, especially since depth is available. }

\section{Overview and contributions}

Our main contribution is a reformulation of the SIMS \cite{qi2018} compositing framework, for the purpose of generating well-aligned RGB images with complementary dense depth (\cref{fig:architecture}).
Therefore, our model enhances SIMS \cite{qi2018} by: 
\begin{enumerate*}
    \item producing \gls{gt} segmentation maps which are better aligned with the RGB image;
    \item synthesizing dense depth from sparse depth alongside RGB images;
    \item using Hu moments as blob descriptors instead of pair-wise \gls{iou} comparisons, in order to speed up blob retrieval.
\end{enumerate*}
As our goal is to build a simple and general framework with minimal handcrafted heuristics, we opt for a simple ordering of object blobs instead of the more complex ordering network that is employed in \cite{qi2018}.

Our work is accompanied by an application of the generated data and \gls{gt} to train semantic segmentation and monocular depth completion tasks.

\section{Related Work}

\reviewaddr{3}{Would it be possible to group listed references? }


\noindent \textbf{Image synthesis}
On one side of the image generation spectrum lies the image compositing process, through which a foreground image or object blob is overlaid on top of a background image or blank canvas.
Although this method is straight-forward and maintains object realism, it often results in artefacts due to differences in appearance of objects coming from different sources (i.e. lighting, shading), or differences in object poses that lead to geometric inconsistencies.
To solve these drawbacks, \cite{martino2016} presents a classic method of pixel-value interpolation while later, \cite{tsai2017} proposes an end-to-end image harmonisation architecture based on \acrshortpl{cnn}.

On the opposite side, lies end-to-end image synthesis, which has been enabled by the rapid advancement of \glspl{gan}~\cite{goodfellow2020}.
Initially unconditional and further conditional on latent representations, segmentation information and style encodings, \glspl{gan} have been successfully applied to image-to-image translation and manipulation, with impressive qualitative results \cite{isola2017,zhu2017,wang2018pix2pixhd,park2019,schonfeld2021}. One of the most common architectures is pix2pixHD \cite{wang2018pix2pixhd} which allows photo-realistic synthesis of $2048\times 1024$ images based on semantic maps, and enables changing of label classes in order to create new scenes. The architecture is further improved in terms of image quality by introducing techniques such as spatially-adaptive normalisation layers \cite{park2019}, segmenting discriminators and 3D-noise sampling to ensure multi-modality \cite{schonfeld2021}.
Although \glspl{gan} are trained to match the distribution of natural images, the realism is still not ideal as the images contain artefacts or lack physical correctness, thus their applicability in robotic downstream tasks is limited.

A different approach to improve robotic tasks makes use of simulators. As a result of the improvements in specialised software and hardware for \gls{cg}, a number of simulators such as GTA \cite{gta}, Carla \cite{carla}, Airsim \cite{airsim}, and Synthia \cite{synthia} have been employed to simulate data from different modalities. While this approach offers access to virtually unlimited data, the process is tedious since it requires creation of environments, assets and scenarios, as well as careful control of parameters. Most disadvantageous, however, is the sim-to-real domain gap, making simulators less than ideal for data generation.

To this end, semi-parametric approaches combine the advantages of model-based and data-driven approaches by taking advantage of the existing data to enforce realism in appearance, and the situational diversity and controllability of simulation. In the context of 2D images, \cite{lee2018} employs two generative models that learn the shape of an object and a plausible, context-aware location in the segmentation map. For videos, GeoSim \cite{chen2021} performs geometry-aware image composition in which new urban scenarios are synthesised by adding dynamic objects (from a pre-built asset bank) on top of existing images, and further blending them in.

\reviewaddr{2}{putting the variable names in Fig. 3 to the images could greatly improve readability. }

\begin{figure*}[t]
\vspace{10pt}
\centering
     \begin{subfigure}[b]{0.24\textwidth}
         \centering
         \includegraphics[width=\textwidth]{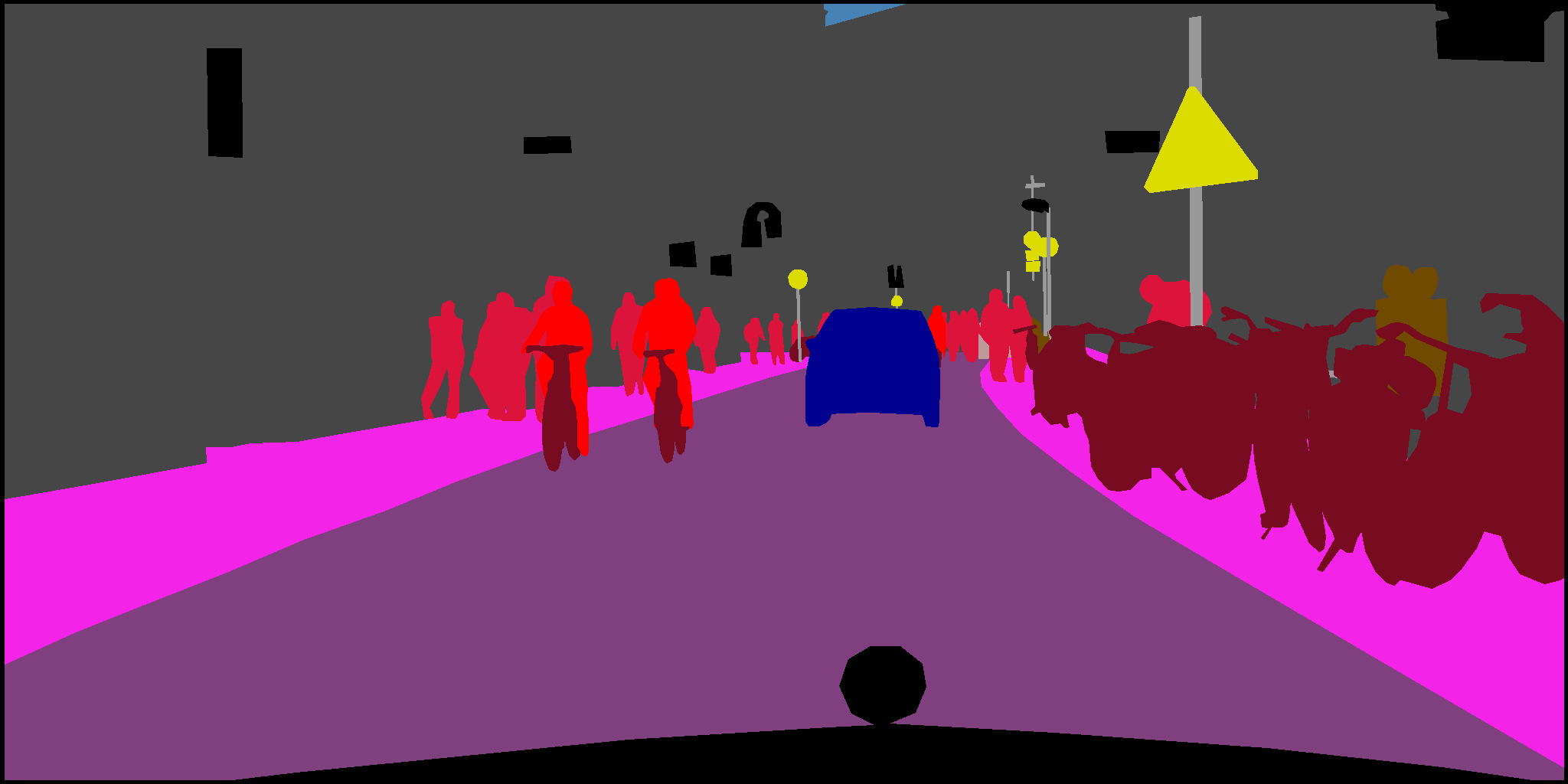}
         \caption{Guide semantic layout $S_{guide}$}
     \end{subfigure}
     \begin{subfigure}[b]{0.24\textwidth}
         \centering
         \includegraphics[width=\textwidth]{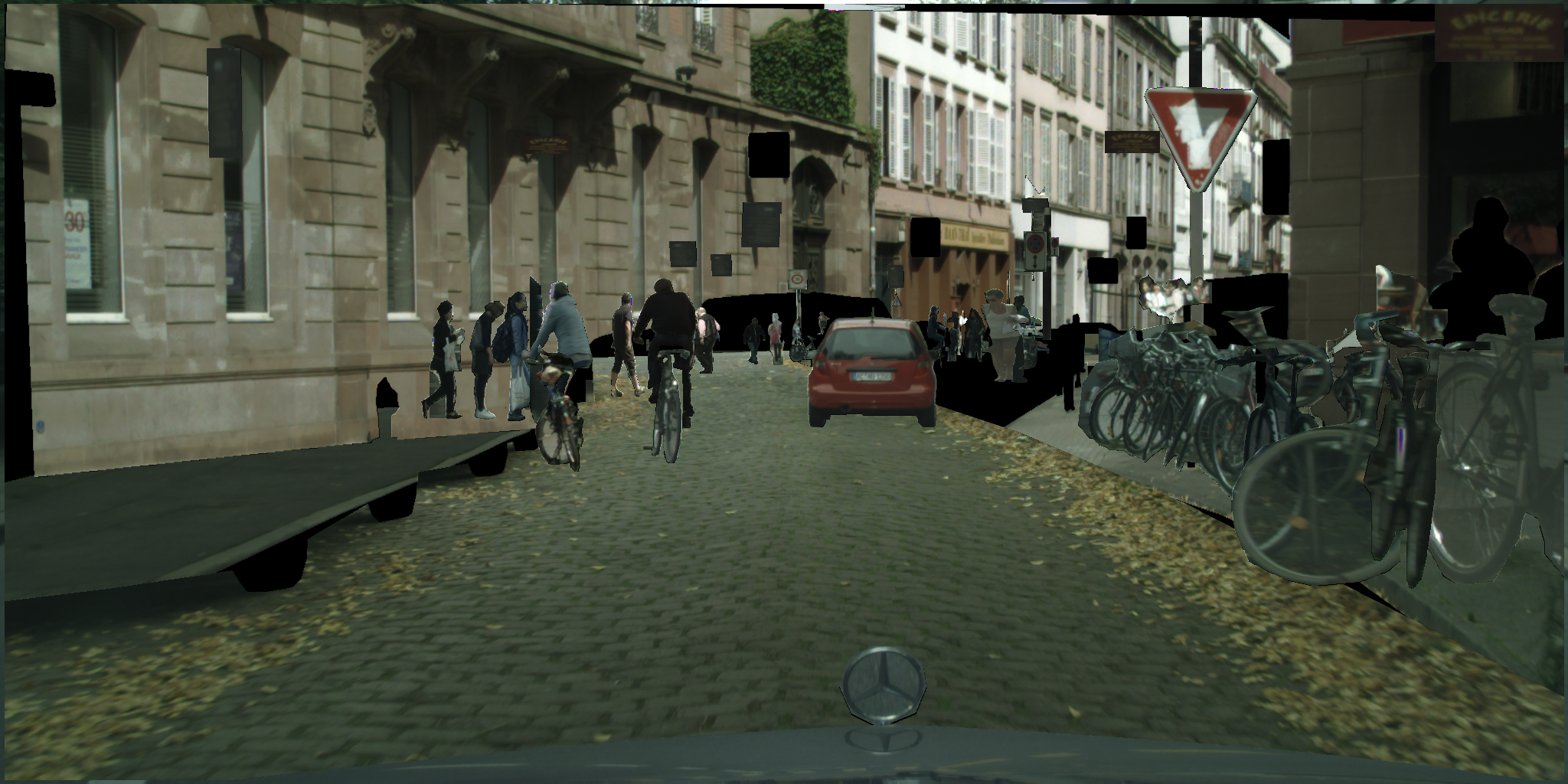}
         \caption{RGB w/ holes $C_{RGB}$}
     \end{subfigure}
     \begin{subfigure}[b]{0.24\textwidth}
         \centering
         \includegraphics[width=\textwidth]{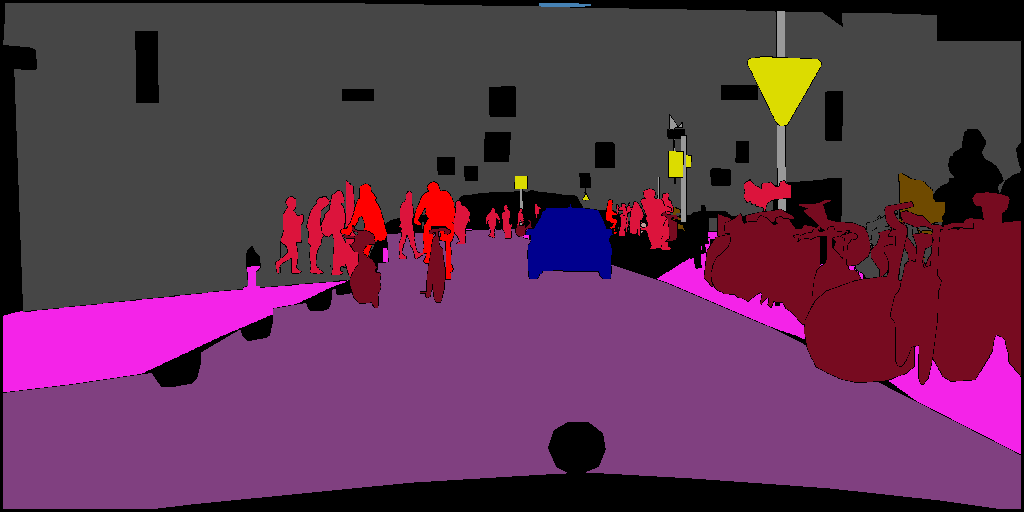}
         \caption{Semantic w/ hole $S_{aligned}$}
     \end{subfigure}
     \begin{subfigure}[b]{0.24\textwidth}
         \centering
         \includegraphics[width=\textwidth]{my_figures/examples_canvasing/munster_000140_000019_gtFine_colorIds.png}
         \caption{Semantic w/out hole $S_{aligned}$}
     \end{subfigure}
     
    \begin{subfigure}[b]{0.24\textwidth}
         \centering
         \includegraphics[width=\textwidth]{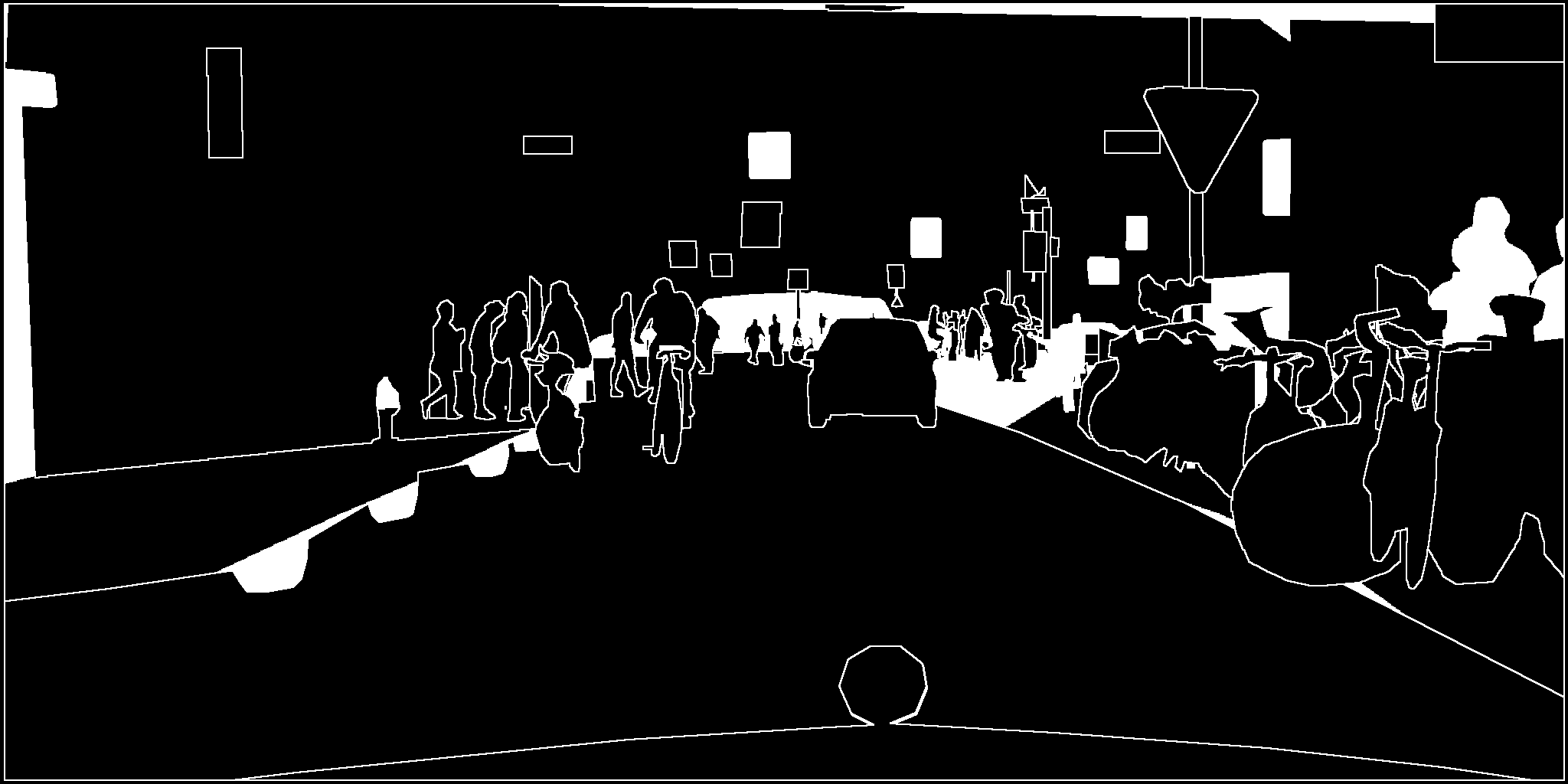}
         \caption{Hole-and-boundary map $H^{\textcolor{white}{*}}$}
     \end{subfigure}
     \begin{subfigure}[b]{0.24\textwidth}
         \centering
         \includegraphics[width=\textwidth]{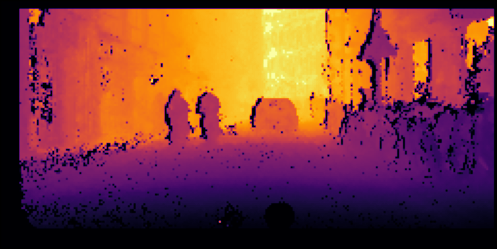}
         \caption{Guiding depth $D_{guide}^{\textcolor{white}{*}}$}
     \end{subfigure}
          \begin{subfigure}[b]{0.24\textwidth}
         \centering
         \includegraphics[width=\textwidth]{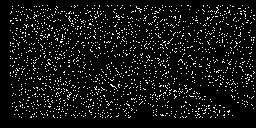}
         \caption{Sparse depth mask $M^{valid}_{depth}$}
     \end{subfigure}
     \begin{subfigure}[b]{0.24\textwidth}
         \centering
         \includegraphics[width=\textwidth]{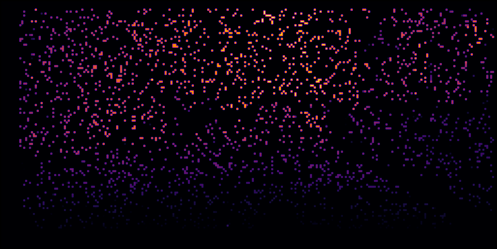}
         \caption{Aligned valid depth $D^{sparse}_{aligned}$}
     \end{subfigure}
    \caption{\small An overview of our canvasing procedure, described in more detail in~\cref{sec:canvasing}.
    The semantic layout $S_{guide}$, shown in~(a) is used to guide the retrieval and placement of blobs onto the RGB canvas $C_{RGB}$, shown in~(b). The associated semantic footprints of the blobs are simultaneously pasted onto a corresponding semantic canvas $S_{aligned}$, shown in~(c), which is further corrected by repairing holes due to mismatched target and original shapes, shown in~(d). The missing areas in $C_{RGB}$ are reflected in a hole-and-boundary map in~(e), which is used to guide the in-painting network later on. The guiding depth in~(f) (which corresponds to $S_{guide}$) is corrected by checking the consistency in class contents between $S_{aligned}$ and $S_{guide}$. Based on valid depth locations and a sampling probability, a sparse depth mask (shown in~(g)) is created, based on which $D_{guide}$ is sparsified, resulting in an aligned sparse depth map, shown in~(h).}
    \label{fig:canvasing}
    \vspace{-0.6cm}
\end{figure*}

\noindent \textbf{In-painting and Harmonization}
Although image compositing is a powerful synthetic data generation tool, the composed canvases are generally affected by missing regions or rough edges.
To this end, image editing tasks such as image in-painting and completion have been used to fill in holes and smooth edges in a perceptually pleasing manner. For example, the architecture in EdgeConnect \cite{nazeri2019} successfully fills in missing regions with fine details, using two modules: a generator that first hallucinates missing edges of objects in the image, and a subsequent generator that fills in pixel values based on the output from the edge network.

In an autonomous-driving context, \cite{Liao2020} makes use of depth information in order to guide a video in-painting task. The authors build a 3D map of frame-wise point clouds which are then projected onto frames, to generate a dense depth map that is further used to sample pixel colors. When removing a dynamic object from the scene, the area is then in-painted using pixel colors from adjacent frames. In contrast, \cite{chen2021} adds dynamic images on top of videos and further smooth the canvas using an in-painting \gls{gan}~\cite{yu2019}.

\noindent \textbf{Depth estimation}
Geometry-based methods include \gls{sfm} where a sequence of 2D images is used to estimate 3D structures via feature matching. This approach is heavily reliant on feature matches which themselves rely on high quality image sequences \cite{Zhao2020}. On the other hand, stereo vision matching requires two viewpoints of the same scene to estimate the 3D structure via disparity maps. Both approaches include either image pairs or image sequences and are thus not applicable for our current setup where we compose a canvas from a single viewpoint.

Sensor-based methods (RGB-D cameras, LiDAR) are capable of retrieving pixel-level depth information but have a limited range and are affected by weather conditions, whereas monocular camera depth estimation has become more popular due to lower weight and cost requirements.

Finally, due to the rapid advancement of deep learning, \acrshort{cnn}-based methods have dramatically improved the accuracy of monocular depth estimation \cite{Godard2017,Alhashim2018,Pillai2019,Lee2019}. \gls{sota} architectures such as PENet \cite{hu2020penet} make use of both image and sparse depth information in a two-branch approach. One branch is designed to exploit color information from the image and sparse depth via an encoder-decoder network with skip connections, while a second depth-dominant branch is designed to output dense depth from a sparse depth map and color information that is processed by the color-dominant branch. 


\section{Methodology}

\reviewaddr{3}{I couldn't see a reference to Fig. 4 anywhere. I think it's better to reference all figures in the text somewhere. Section IIA seems like a good place to provide Fig. 4 as a supporting example, despite not appearing until a few pages later. }

We aim to synthesise an image and its corresponding pixel-wise segmentation and dense depth map, from initial guiding semantic layout and sparse depth, and a database of objects. The process is split into four steps: 
\begin{enumerate*}
    \item the creation of an object blob database from a source set,
    \item a canvasing step in which a series of compositing rules are used to position objects retrieved from the database onto a blank image canvas and a corresponding segmentation canvas, based on a guiding segmentation layout,
    \item a sparse depth map composition step, where any existing depth cues are used to create a sparse depth map that corresponds to the newly created image canvas,
    \item an in-painting step in which the image canvas is harmonised and the sparse depth map is densified.
\end{enumerate*}
A graphical representation of the system can be found in \cref{fig:architecture}.

\subsection{Object blob database}

Let $N$ be an image dataset where for each image $I_i \in N$, object instances have been labelled through segmentation and instance masks $S_i$ and classified as one of $L$ classes.
Our first objective is to decompose $N$ into a dataset $B$ of object blobs, where each class of objects is stored separately.
More specifically, for each blob $b_j$ in each image and segmentation pair $(I_{i},S_{i})\in N_{train}$ from the train set, we extract its corresponding binary mask $m_j$ and label $l_j$; where $b_j\in \mathbb{R}^{h_{j}\times w_{j}\times 3}$, $m_j\in {\{0,1\}}^{h_{j}\times w_{j}\times 1}$ and $l_{j}\in L$, where $h_{j}\times w_{j}$ represent the dimension of the tight bounding rectangle inscribing the blob.
Empirically, we ignore object blobs whose areas are smaller than $1000$ pixels, and prefer larger, high-quality blobs that can then be resized.
For object classes lacking instance segmentation, we attempt to obtain them by performing \textit{connected components} \cite{connected_components}.

Alongside each blob $b_{j}$, mask $m_j$, and label $l_j$ from image $I_{i}$, we store its corresponding shape descriptor $q_{j}$.
The original framework of \cite{qi2018} uses \gls{miou} as a metric for blob matching however, this is difficult to scale to large numbers of blobs as any retrieval requires pair-wise comparisons over the entire dataset that belong to a particular class.
To overcome this limitation, we use Hu moments~\cite{hu_moments} shape descriptors, which -- beyond being invariant to translation, scale, rotation, and reflections -- can be easily stored in a database and quickly compared, allowing faster retrieval. 

To keep our assessment in line with previous works, we focus on the Cityscapes \cite{cityscapes} dataset as a source of object blobs.
However, any other dataset which provides instance segmentation can be used.

\subsection{Canvasing\label{sec:canvasing}}
Elements of our canvasing procedure are shown in~\cref{fig:canvasing}.
In this second stage we compose both an image canvas $C_{RGB}\in \mathbb{R}^{w\times h \times 3}$ based on a guiding semantic layout $S_{guide}$, and a corresponding segmentation map $S_{aligned}\in \mathbb{R}^{w\times h \times 1}$ that is aligned to $C_{RGB}$.

The canvases $C_{RGB}$ and $S_{aligned}$ are composed starting from blank canvases onto which objects from database $B$ and their masks are pasted. Based on a guiding semantic layout $S_{guide}$ that the compositing should follow, we retrieve blobs that have a similar shape to that of the object's footprints from the guiding segmentation layout $S_{guide}$. More specifically, for a segmentation map $S_{guide}^{i}\in N_{val}$ from the validation set, shown in~\cref{fig:canvasing}(a), for each instance label with corresponding mask $(l_{j},m_{j})\in S_{guide}^{i}$, we compute a Hu moments shape descriptor and use it to find the closest blob $b_j\in B$ and paste it on the canvas $C_{RGB}$, shown in~\cref{fig:canvasing}(b). Similarly, we paste the mask of the blob $b_j$ onto the semantic canvas $S_{aligned}$, shown in~\cref{fig:canvasing}(c). Due to invariance to reflection of the shape descriptor based on Hu moments, we test the suitability of both orientations of the retrieved blob by flipping it horizontally and comparing the \gls{iou}s with the target footprint. Vertical flipping can also be used, but for most classes it is not applicable.
As a general rule, we first place sky, buildings, road and pavement blobs, followed by the rest of the static objects and finally, the dynamic ones. 

As there is often a mismatch between the target and original shapes, the RGB canvas $C_{RGB}$ and the semantic canvas $S_{aligned}$ may contain holes. The RGB canvas holes will be in-painted by the in-painting model in a subsequent step using a hole-and-boundary map described below. The semantic canvas holes are repaired using a simple strategy:
\begin{enumerate}
    \item For any holes that correspond to static classes in the original guiding semantic map $S_{guide}^{i}$, we simply copy over the semantic information from $S_{guide}^{i}$;
    \item For any remaining holes that do not correspond to static classes in the original guiding semantic map $S_{guide}^{i}$, we create a copy of the semantic canvas $S_{aligned}^{copy}$, dilate the static class footprints in this copy until no more holes exist, and finally transfer information corresponding to the location of the initial holes into $S_{aligned}$.
\end{enumerate}

\reviewaddr{1}{ "In contrast to the method used by SIMS we produce a segmentation map that is better aligned with the output RGB canvas". Which exp results do confirm that statement? What metric is used? }

This strategy allows us to keep blobs corresponding to dynamic classes unmodified, but also assumes that static classes (road, sky, pavement, vegetation etc) are mainly texture-based and can be easily in-painted by the in-painting model.
An example of the results of the hole-filling strategy can be seen in \cref{fig:canvasing}(d). 

Finally, a hole-and-boundary map $H$ (\cref{fig:canvasing}(e)) that encodes the location of any remaining holes in the RGB canvas along with the boundaries of all pasted blobs will be used as guide in the in-painting phase.

\subsection{Sparse depth}
In the third stage, we optionally create a sparse depth map. For a guiding semantic layout $S_{guide}$ that has an available depth map $D_{guide}$ (in~\cref{fig:canvasing}(f)), we can also create a sparse depth map $D_{aligned}^{sparse}$ that corresponds to the newly generated RGB canvas $C_{RGB}$ and its segmentation map $S_{aligned}$, as shown in~\cref{fig:canvasing}(g).

In order to produce the sparse depth map $D_{aligned}^{sparse}$, we first compute the intersection between the original segmentation guiding layout $S_{guide}$ and the segmentation map of the RGB canvas $S_{aligned}$, yielding a validity mask $M_{depth}^{valid}$, with the assumption being that only points from $S_{aligned}$ that have kept the same class as in $S_{guide}$ are valid as a source of depth information. We then sample locations from the validity mask $M_{depth}^{valid}$, and produce a sparse depth map $D_{aligned}^{sparse}$ with depth information taken from the depth map $D_{guide}$ of the guiding semantic layout $S_{aligned}$. The choice of a sparse depth map implies that both dense and sparse sources of depth (e.g. LIDAR) can thus be used with our method.

\subsection{Image in-painting and depth synthesis}

In the fourth and final stage we train a \gls{gan}-based in-painter to both synthesise a final RGB image by in-painting holes in the RGB canvas $C_{RGB}$ and to densify the associated sparse depth map $D_{aligned}^{sparse}$.
For this, we choose \gls{sota} components for the pipeline, such as a SPADE-based in-painting generator \cite{park2019}, and a segmenting discriminator based on OASIS \cite{schonfeld2021}. The overall architecture is shown in \cref{fig:architecture}.

The in-painting generator takes as input the RGB canvas $C_{RGB}$, hole-and-boundary map $H$, semantic canvas $S_{aligned}$, edge map $E$ (encoding the edges of blob instances, generated from the instance map), sparse depth map $D_{aligned}^{sparse}$ and sparse depth mask $M_{aligned}^{sparse}$ that indicates which depth values are valid. The in-painter then outputs both an in-painted RGB image $I_{RGB}$ with no holes and a dense depth map $D_{aligned}^{dense}$ corresponding to the in-painted RGB image.

An intuitive explanation of the low \gls{fid} scores (indicative of high image quality) of compositing approaches such as \cite{qi2018} is the fact that most of the areas in the generated images come from natural or real blobs. As such, our implementation also attempts to use as much information from the original RGB canvas as possible. Specifically, in our case, this means replacing blobs from the in-painted RGB output with blobs from the input RGB canvas using the hole-and-boundary mask to create a final RGB in-painted image. However, initial training experiments failed to converge, resulting in blurry in-painted regions, a possible explanation for this being the sparse gradients. The framework was therefore modified to alternatively pass both the raw output from the generator and the final in-painted and blended image to the discriminator. 

\reviewaddr{1}{Several concepts in the equations 1,3 and 4 were not introduced: $t_{i,j,c}, E_t, D_{GT}$ }

The segmentation-based discriminator outputs a $L+1$ one-hot encoded prediction of the class of each pixel in the image. Similar to \cite{schonfeld2021}, we have $L$ real classes and $1$ fake class.  We use a cross-entropy loss between the one-hot encoded prediction and the \gls{gt} one-hot encoded segmentation map. Given real images $x_R$ and canvas images $x_C$, the discriminator loss is:
\hspace{-5cm}
\begin{equation}
\small{
\label{eq:loss_discr}
\begin{split}
\hspace{-5cm}
     \mathcal{L}_{D}\!=\!-\mathbb{E}_{(x_R)}\!\left[\!\sum_{c=1}^{L}\!\alpha_{c}\!\sum_{i, j}^{H \times W}\!t_{i, j, c}\!\log D(x_R)_{i, j, c}\!\right]\!\\\!-\mathbb{E}_{(x_C)}\!\left[\!\sum_{i, j}^{H \times W}\!\log D(G(x_C))_{i, j, c=L+1}\!\right]\!
\end{split}
}%
\end{equation}
where $i,j$ represent the pixel coordinates and $c$ represents the channel.
In this equation, $t_{i, j, c} = 1$ if the pixel $(i, j)$ belongs to the class $c$ and $0$ otherwise.
The goal is for the discriminator to output a value of $1$ at pixel location $(i,j)$ in the corresponding channel of a particular class when the input is a real image, such that the first term $\rightarrow 0$. On the contrary, when the input to the discriminator is an in-painted canvas, the goal is for the discriminator to output a value of $1$ in the $(N+1)$-th channel, thus signalling that the pixel is fake, minimising the second term. 

The generator tries to minimize the following loss:
\hspace{-5cm}
\begin{equation}
\label{eq:loss_gen}
\small{
\mathcal{L}_{G}=
-\mathbb{E}_{(x_C)}\!\!\!\left[\!\sum_{c=1}^{L}\!\alpha_{c}\!\!\!\sum_{i, j}^{H \times W}\!\!\!t_{i, j, c}\!\log D(G(x_C))_{i, j, c}\!\right]\!
}
\end{equation}
From the point of view of the generator, the goal is to fool the discriminator, i.e. to output a value of $1$ in the corresponding channel of a particular class when the input is a fake image, such that the term $\rightarrow 0$.

The weights $\alpha_{c} \in \mathbb{R}$ in \cref{eq:loss_discr,eq:loss_gen} are used to weigh the loss according to the class balance of the example, with 
\begin{equation}
\small{
\alpha_{c}=\frac{H \times W}{\sum_{i, j}^{H \times W} t_{i, j, c}}
}
\end{equation}


Finally, for the depth densification task, we use a \gls{mse} loss between the predicted dense depth $D^{dense}_{aligned}$ and the \gls{gt} depth $D^{dense}_{GT}$:
\begin{equation}
    \mathcal{L}_{depth}\!=\!\|(D^{dense}_{aligned} - D^{dense}_{GT})^2\|
\end{equation}
The final combined loss is:
$
    \mathcal{L}_{total}\!=\!\mathcal{L}_{D} + \mathcal{L}_{G} + \lambda_d\mathcal{L}_{depth}
$
where $\lambda_d$ controls the strength of the \gls{mse} loss in relation to the other two terms.
\vspace{-0.025cm}
\subsection{Training details}

\reviewaddr{Gadd}{unclear to me -perhaps useful to state what loss we use to (pre-)train space in-painter for RGB?}

In order to train the in-painting generator, we cannot directly use the outputs of our Canvaser, as a corresponding \gls{gt} image and depth do not exist. Instead, starting from the original Cityscapes train set \cite{cityscapes}, we create a dataset of eroded RGB images and corresponding segmentation and sparse depth maps to emulate a canvas. For each input RGB image, we make use of its instance map to generate a boundary map, to which we add randomly-shaped masking polygons to simulate missing areas. This hole-and-boundary map $H$ is further used to mask the RGB image. Such a step is necessary to mimic the canvasing process that would take place at inference time. To increase the generalisation ability, we dilate sections of map $H$ using randomly-sized kernels.

\reviewaddr{2}{ In Fig. 2 and Fig. 5, some text is barely readable in the printed version }

\reviewaddr{2}{Using the variable names (e.g. $x_R$, $x_C^{inpainted}$) more consistently in the graphics (Fig. 2, Fig. 5) would be helpful. }

The depth map is generated from the disparity image and then uniformly sampled with probability $p_{sample}$ to generate a sparse depth map and a corresponding sparse depth mask. Finally, the edge map is created from the segmentation and instance maps.

\vspace{-0.2cm}
\section{Evaluation}

\reviewaddr{1}{Contradiction on the segmentation model used for the experiments.  }

All image synthesis models are trained with the Adam \cite{adamoptim} optimizer for $50$ epochs and a learning rate of $0.01$. For our model, we set $\lambda_d$ to $100$.

\subsection{Model evaluation}

In line with previous work, we compare our model in terms of perceptual quality and alignment of synthesised images with their \gls{gt} \cite{schonfeld2021}. We choose strong \gls{sota} baselines, such as pix2pixHD \cite{wang2018pix2pixhd}, SPADE \cite{park2019}, SPADE+, and OASIS \cite{schonfeld2021} for the parametric approach and SIMS \cite{qi2018} for the semi-parametric approach.
In measuring perceptual quality, we use \gls{fid}~\cite{FID}, as it has been shown to be in line with human judgements.
In measuring semantic alignment between generated data and \gls{gt} labels, we employ a DRN-D-105 \cite{drn} network -- pretrained for the task of semantic segmentation -- and apply it to the generated images to measure the \gls{miou} between the resulting segmentation and the \gls{gt} segmentation masks, as in \cite{schonfeld2021}.
The model evaluation results are reported on the validation set of Cityscapes \cite{cityscapes}. 

\reviewaddr{2}{For the image data evaluation (IV.B), how is the exact evaluation regime? Which segmentation mask is taken from which dataset and how exactly is the evaluation performed? }

\subsection{Image data evaluation}\label{img_data_eval_section}
To test the usefulness of the synthesised image data as training data, we employ DRN-D-22 \cite{drn} as it is faster to train.
We train our model, SIMS and OASIS on the Cityscapes train set, and evaluate them on the validation set to produce training data at resolution $512\!\times\!256$ \footnote{experiment equivalent to those reported in \cite{qi2018} ``Cityscapes-fine''}.
We then train 3 individual instances of DRN-D-22 on the output of each of the models in the same training regime as above, and test them on the out-of-domain A2D2 dataset \cite{a2d2}. We report \gls{miou} results. 

\subsection{Depth evaluation}
\begin{figure}[htbp]
    \centering
  \includegraphics[scale=0.55, trim=24.0cm 9.0cm 29.0cm 12.0cm]{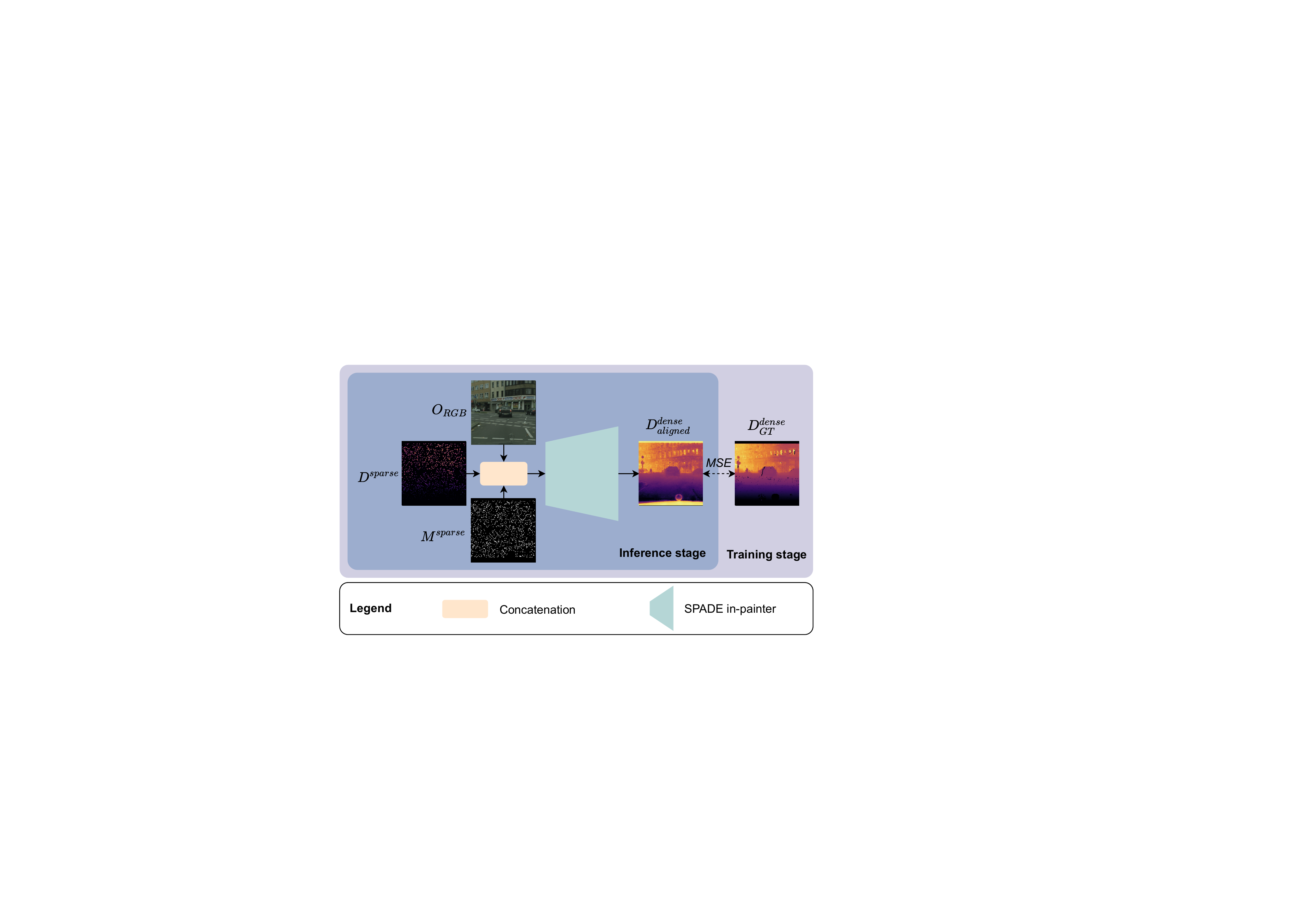}
  \caption{\small We employ a light-weight depth-completion model based on a SPADE generator and use \gls{mse} loss for training.\reviewaddr{Gadd}{Is this a different set of weights to RGB SPADE?}
 }
  \label{fig:depth_arch}
    \vspace{-7pt}
\end{figure}

To test the quality of our estimated dense depth associated with the in-painted images, we employ a light-weight depth-completion model, based on the architecture from \cref{fig:architecture}.
The network only takes as input the concatenation of an RGB image, sparse depth map and corresponding depth mask and outputs predicted dense depth. We make use of \gls{mse} loss between the aligned predicted dense depth $D^{dense}_{aligned}$ and \gls{gt} dense depth $D^{dense}_{GT}$, as illustrated in \cref{fig:depth_arch}.

We train 3 individual instances of our monocular depth network on the generated RGB images of SIMS, SPADE and our model, and evaluate them out-of-domain, on the validation split of the KITTI dataset \cite{kitti_depth}. Since the depth estimator is supervised with dense depth, we use the synthesised depth for our model as target, and the \gls{gt} depth of Cityscapes for SIMS and SPADE, as they only synthesise RGB images. We report \gls{rmse} results.

\begin{figure*}[t]
\centering
     \begin{subfigure}[b]{0.31\textwidth}
         \centering
         \figuretitle{\small OASIS}
         \includegraphics[width=\textwidth]{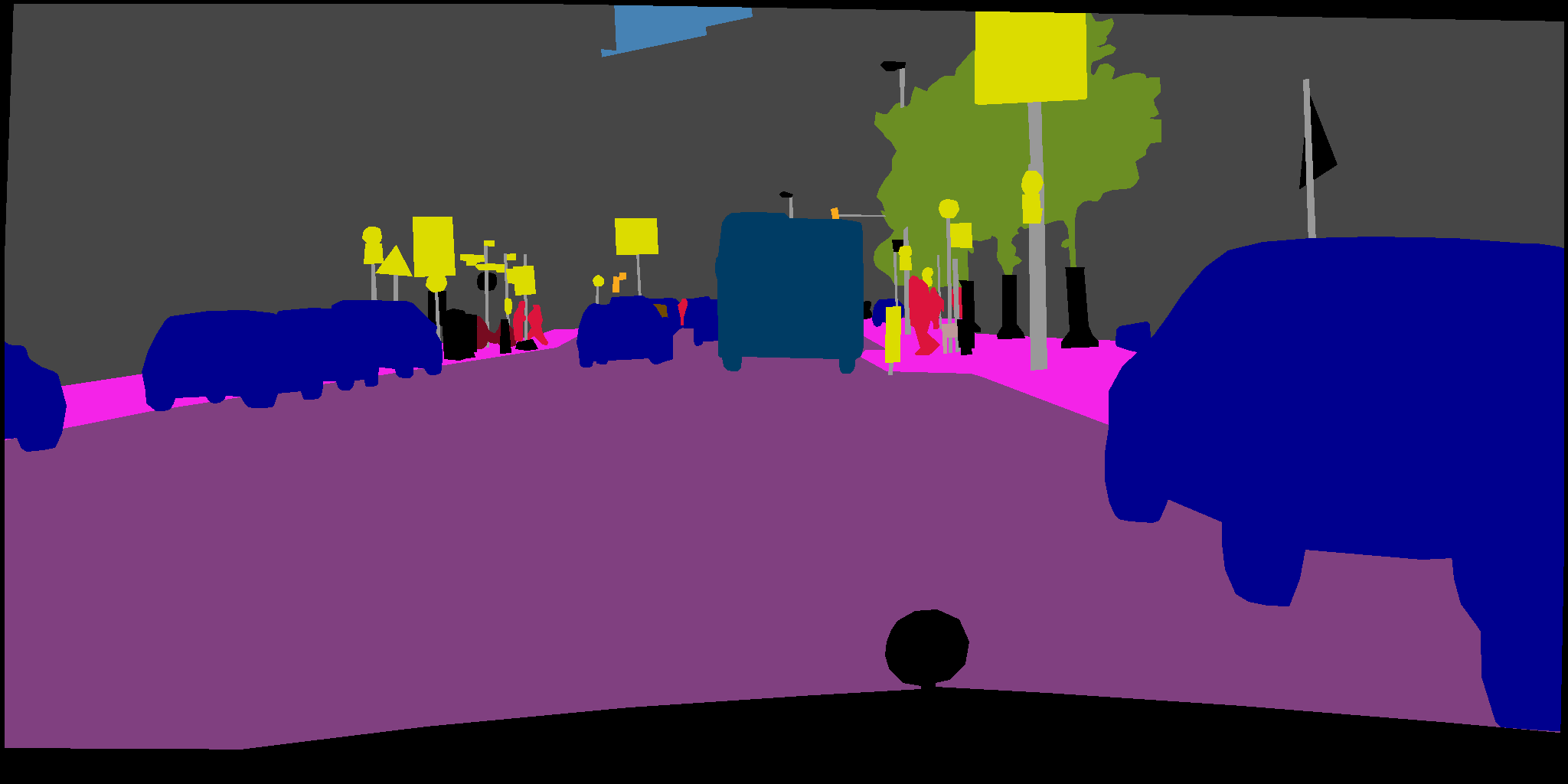}
     \end{subfigure}
     \vspace{0.1cm}
     \begin{subfigure}[b]{0.31\textwidth}
         \centering
         \figuretitle{\small SIMS}
         \includegraphics[width=\textwidth]{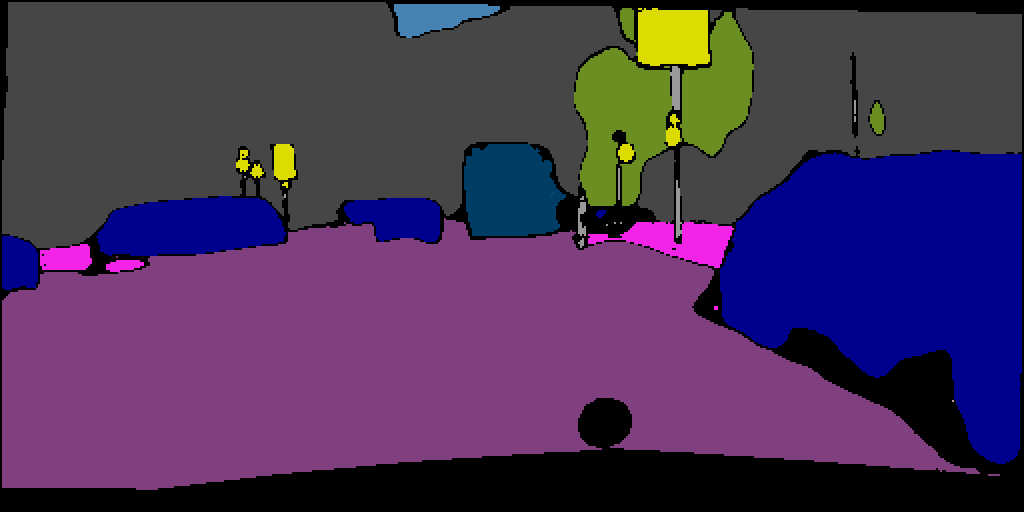}
     \end{subfigure}
     \begin{subfigure}[b]{0.31\textwidth}
         \centering
         \figuretitle{\small OURS}
         \includegraphics[width=\textwidth]{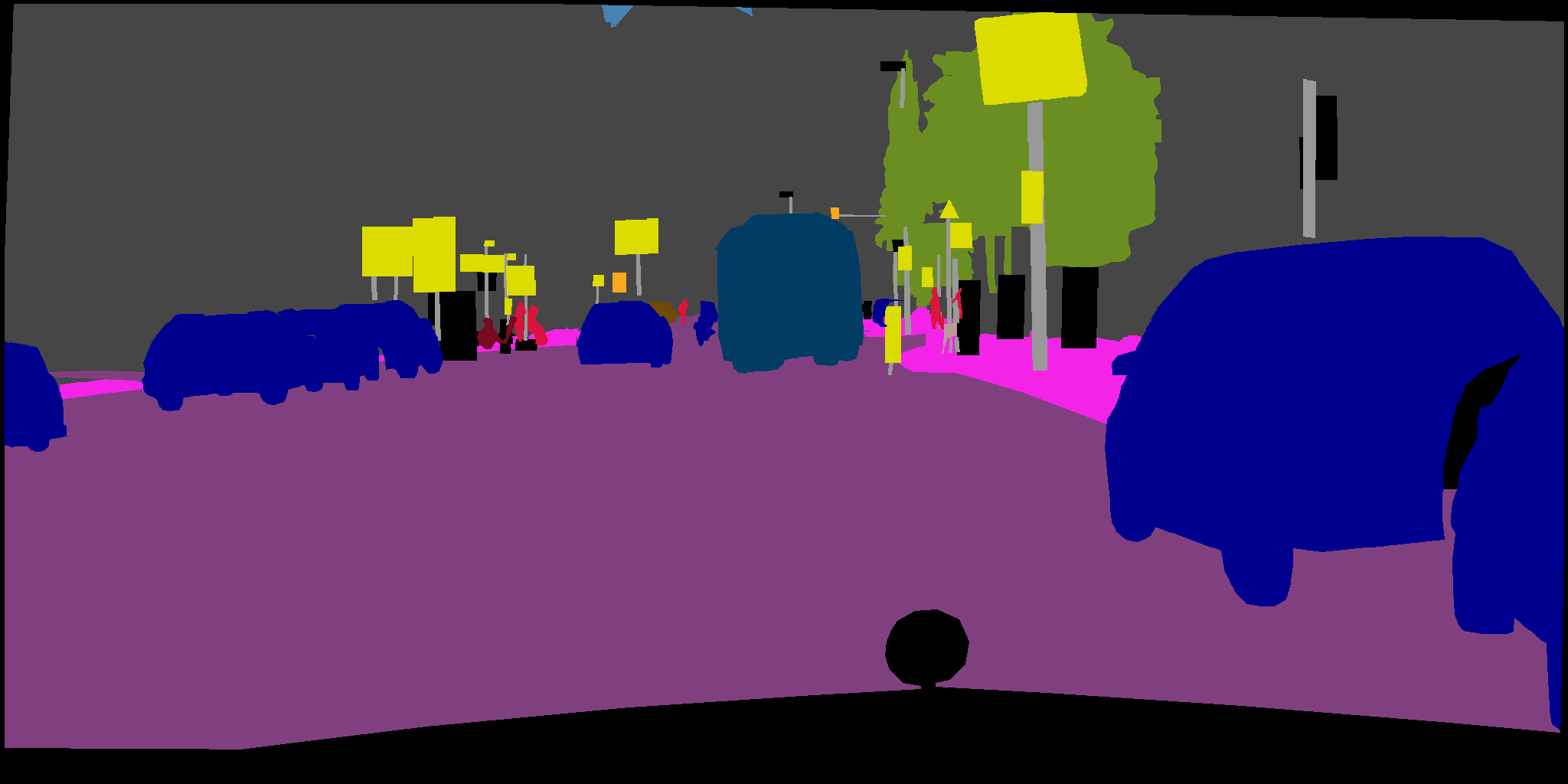}
     \end{subfigure}    

     \begin{subfigure}[b]{0.31\textwidth}
         \centering
         \includegraphics[width=\textwidth]{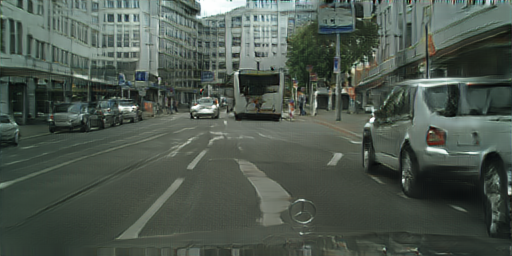}
     \end{subfigure}
     \vspace{0.1cm}
     \begin{subfigure}[b]{0.31\textwidth}
         \centering
         \includegraphics[width=\textwidth]{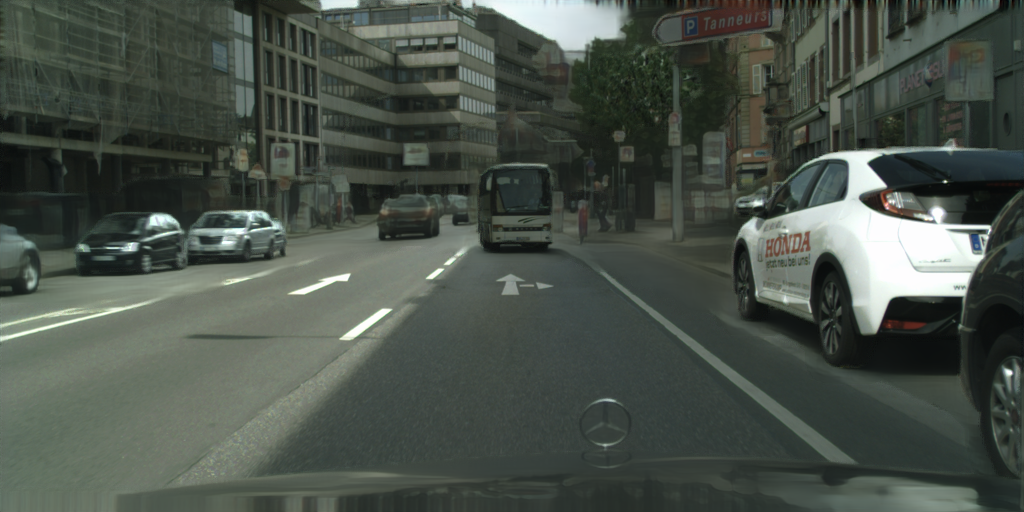}
     \end{subfigure}
     \begin{subfigure}[b]{0.31\textwidth}
         \centering
         \includegraphics[width=\textwidth]{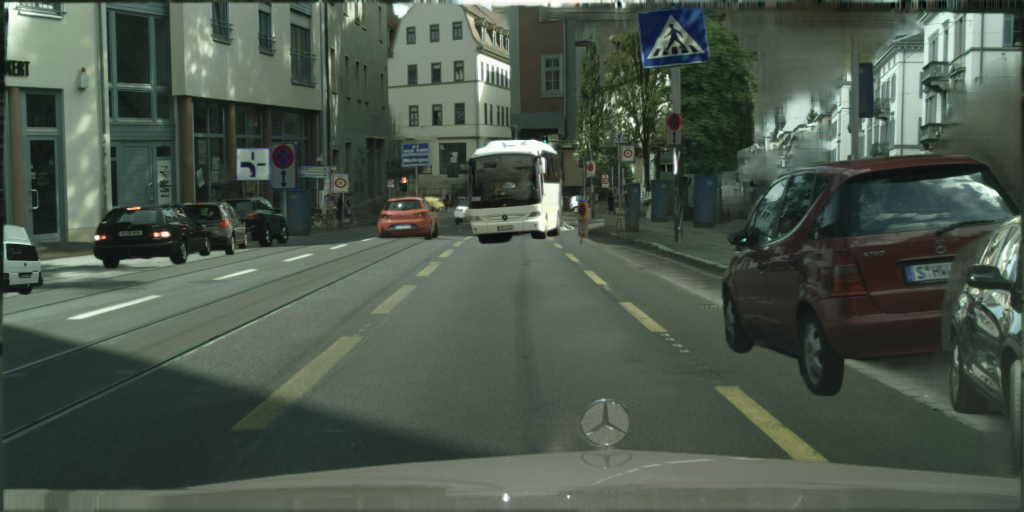}
     \end{subfigure}
    \caption{\small RGB images and corresponding \gls{gt} segmentation for OASIS, SIMS and our method. We note that OASIS has aligned semantic segmentation but lower image quality, SIMS has high image quality but imperfectly-aligned segmentation, while our method has both good image quality and well-aligned segmentation.}
    \label{fig:generated_data}
    \vspace{-20pt}
\end{figure*}

\section{Results}
\label{sec:results}

\reviewaddr{1}{Qualitative evaluation is reduced to oly ione sample and 3 benchmarks which might be insufficient. }

\reviewaddr{1}{ Although qualitative image generation results are not convincing, performance of the proposed method on the downstream task is significantly better. Hence several concerns.}

\reviewaddr{1}{ Artifacts claimed to be tackled by the method in II.A (lighting, shading, poses etc) still persist in the resulting images - shades below cars, inconsistent orientation of the bus, inconsistent lane markings, spatial inconsistency of traffic sign, bus lighting vs car lighting }

\review{2}{ Fig. 6 really suffered from image compressio }

\review{2}{The arrows in Fig. 6 could simply be aligned with the axes. }

\begin{figure}[!h]
    \centering
    \vspace{.3cm}
    \includegraphics[scale=0.4,trim={0.8cm 0.5cm 0.3cm 0.6cm},clip]{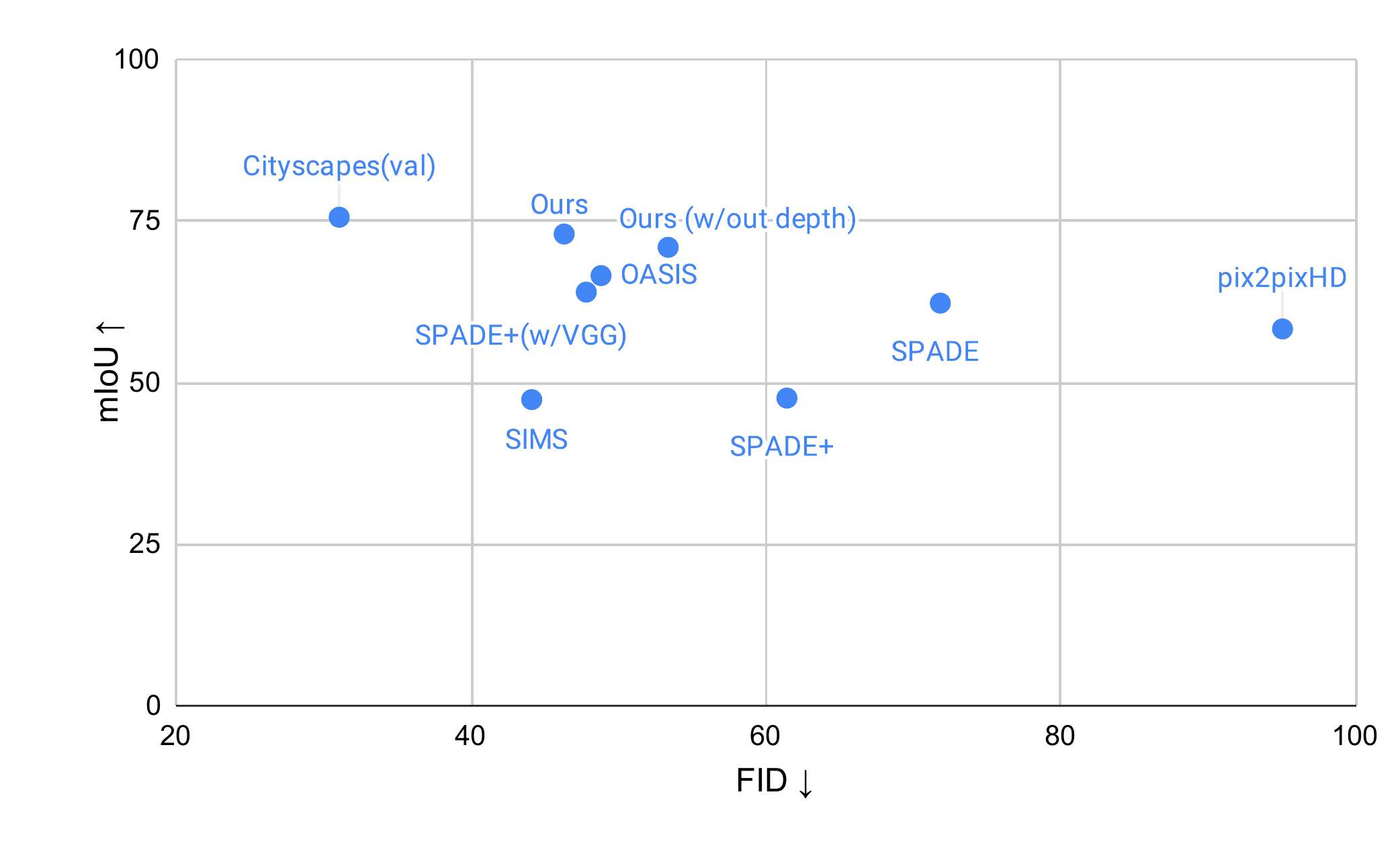}
    \vspace{-0.3cm}
    \caption{\small Our model scores the highest in terms of \gls{miou}, with a highly competitive \gls{fid}. The performance of the real Cityscapes validation split is shown as an upper bound.}
    \label{fig:miou_fid}
    \vspace{-20pt}
\end{figure}

\reviewaddr{2}{However, as a human, SIMS results are much more convincing. This makes me wonder how the proposed approach would perform in a perceptual experiment as conducted for SIMS. }

\Cref{tab:model_benchmark} presents results from benchmarking the data generated by the model itself. Images produced by our method are better aligned with the \gls{gt} segmentation, yielding a significantly higher \gls{miou} score than all previous methods, approaching the performance of the real Cityscapes validation split. Additionally, it should be noted that, as with other methods that make use of a pretrained segmentation network, the resulting \gls{miou} is also dependent on how closely the distribution of generated images follows the distribution of the dataset on which the segmentation network was trained. Although both SIMS and our method aim to preserve as much real blob components as possible, the higher \gls{miou} indicates that our method benefits from better alignment, at the cost of a slight increase in \gls{fid}, which nevertheless remains highly competitive, indicating that our method achieves a higher perceptual quality compared to parametric methods. We further conduct an ablation study to determine the usefulness of the depth completion component of our in-painter. Our approach benefits from the addition of depth completion, which both increases the \gls{miou} by 2.1 percentage points and decreases \gls{fid} by 7 points.
Qualitative examples of image synthesis can be seen in \cref{fig:generated_data} and a visual comparison of \gls{sota} methods in \cref{fig:miou_fid}.

\Cref{tab:train_data_benchmark} presents results from benchmarking the 2 downstream tasks trained using the data generated by the 3 models.
We emphasise that this benchmarking is meant to rank the performance of synthetic training data rather than outperform \gls{sota} segmentation and depth completion architectures.
The segmentation task trained using data generated by our model produces better results when tested on an out-of-domain test set (A2D2) as compared to training using data produced by SIMS and OASIS.
We interpret these results as a reflection of the importance of good alignment between generated images and segmentation maps when training a new task, as both our method (explicit segmentation alignment) and OASIS (implicit segmentation alignment) outperform SIMS in the segmentation task.
In terms of depth completion, our method outperforms competing methods, when the depth synthesised by our in-painter is used to train the depth completion task. Conversely, and as a mini-ablation study, using the original guiding \gls{gt} depth instead of our synthesised depth leads to worse results, possibly indicating that the synthesised depth is better matched to the generated images, compared to the original guiding \gls{gt} depth.

\begin{table}[]
\renewcommand{\arraystretch}{1.2}
\centering
\begin{tabular}{c|c|cc}
\toprule
Model                   & VGG & FID $\downarrow$  & mIoU $\uparrow$ \\
\hline
\hline
SIMS\tablefootnote{\label{their_exp}Results quoted from \cite{schonfeld2021} (Table 1, page 8).}                    & \textcolor{mygreen}{\cmark} & 49.7 & 47.2 \\
SIMS\tablefootnote{\label{our_exp}Results of our own experiments, with the official implementation}                    & \textcolor{mygreen}{\cmark} & \textbf{44.1} & 47.4 \\
pix2pixHD\footref{their_exp}              & \textcolor{mygreen}{\cmark} & 95.0 & 58.3 \\
SPADE\footref{their_exp}                  & \textcolor{mygreen}{\cmark} & 71.8 & 62.3 \\
\hline
\multirow{2}{*}{SPADE+\footref{their_exp}} & \textcolor{mygreen}{\cmark} & 47.8 & 64.0 \\
                        & \textcolor{myred}{\xmark} & 61.4 & 47.6 \\
\hline
OASIS\footref{their_exp}                   & \textcolor{myred}{\xmark} & 47.7 & 69.3 \\
OASIS\footref{our_exp}                   & \textcolor{myred}{\xmark} & 48.8 & 66.6 \\
Ours w/out depth                    & \textcolor{mygreen}{\cmark}   & 53.3    & 70.9 \\
Ours                    & \textcolor{mygreen}{\cmark}   & 46.3    & \textbf{73.0} \\   
\hline
Cityscapes (val set)                    & -   & 31.05    & 75.6 \\   
\bottomrule
\end{tabular}
\caption{\small Model benchmark by alignment and perceptual quality.}
\label{tab:model_benchmark}
\vspace{-.35cm}
\end{table}

\reviewaddr{2}{How would the proposed approach perform in the experiment depicted in Table 2 in the SIMS paper. Why did the authors deviate from this setup? }

\begin{table}[]
\renewcommand{\arraystretch}{1.2}
\centering
\begin{tabular}{c|cc}
\toprule
Model & mIoU $\uparrow$  & RMSE (m) $\downarrow$ \\
\hline
\hline
OASIS w/ GT depth & 28.31 & 9.34    \\
SIMS w/ GT depth  & 24.36 & 4.48    \\
Ours w/ synthesised depth  & \textbf{29.87} & \textbf{3.90}   \\
Ours w/ GT depth & - & 5.99   \\
\bottomrule
\end{tabular}
\caption{\small Train data benchmark on semantic segmentation and depth completion.}
\label{tab:train_data_benchmark}
\vspace{-.6cm}
\end{table}

\review{2}{Many visual artifacts seem to stem from scaling blobs. This could be remedied by storing not only the Hu moments, but also depth and retrieving blobs with a trade-off between depth and shape match. With this, e.g. unrealistically large signs could be avoided. Also, it would solve the problem that scaling blobs introduces as depth is not scaled accordingly. E.g. a small, but enlarged vehicle would be estimated closer than a larger, originally sized one, both sharing the same shape. }

\section{Conclusions}

In this paper we have presented a compositing image synthesis method that makes use of a set of simple heuristics to compose not only an RGB image canvas but also a corresponding well-aligned semantic segmentation map and a sparse depth map, followed by an in-painting stage that yields a high quality RGB image and a dense depth map. We successfully benchmark the quality of the data produced by the model, showing an increase of 3.7 percentage points for \gls{miou} over other \gls{sota} parametric and semi-parametric approaches and a highly competitive \gls{fid} score. Additionally, we benchmark the suitability of our synthesised data as training data for semantic segmentation and depth completion, showing that the data produced by our model is more suitable for this purpose than the other methods. 

\vspace{20pt}





\bibliographystyle{IEEEtran}
\bibliography{my_bib}

\begin{thebibliography}{10}
\providecommand{\url}[1]{#1}
\csname url@rmstyle\endcsname
\providecommand{\newblock}{\relax}
\providecommand{\bibinfo}[2]{#2}
\providecommand\BIBentrySTDinterwordspacing{\spaceskip=0pt\relax}
\providecommand\BIBentryALTinterwordstretchfactor{4}
\providecommand\BIBentryALTinterwordspacing{\spaceskip=\fontdimen2\font plus
\BIBentryALTinterwordstretchfactor\fontdimen3\font minus
  \fontdimen4\font\relax}
\providecommand\BIBforeignlanguage[2]{{%
\expandafter\ifx\csname l@#1\endcsname\relax
\typeout{** WARNING: IEEEtran.bst: No hyphenation pattern has been}%
\typeout{** loaded for the language `#1'. Using the pattern for}%
\typeout{** the default language instead.}%
\else
\language=\csname l@#1\endcsname
\fi
#2}}

\bibitem{yu2020bdd100k}
F.~Yu, H.~Chen, X.~Wang, W.~Xian, Y.~Chen, F.~Liu, V.~Madhavan, and T.~Darrell,
  ``Bdd100k: A diverse driving dataset for heterogeneous multitask learning,''
  in \emph{Proceedings of the IEEE/CVF conference on computer vision and
  pattern recognition}, 2020, pp. 2636--2645.

\bibitem{cityscapes}
M.~Cordts, M.~Omran, S.~Ramos, T.~Rehfeld, M.~Enzweiler, R.~Benenson,
  U.~Franke, S.~Roth, and B.~Schiele, ``The cityscapes dataset for semantic
  urban scene understanding,'' in \emph{Proc. of the IEEE Conference on
  Computer Vision and Pattern Recognition (CVPR)}, 2016.

\bibitem{a2d2}
\BIBentryALTinterwordspacing
J.~Geyer, Y.~Kassahun, M.~Mahmudi, X.~Ricou, R.~Durgesh, A.~S. Chung,
  L.~Hauswald, V.~H. Pham, M.~M{\"u}hlegg, S.~Dorn, T.~Fernandez,
  M.~J{\"a}nicke, S.~Mirashi, C.~Savani, M.~Sturm, O.~Vorobiov, M.~Oelker,
  S.~Garreis, and P.~Schuberth, ``{A2D2: Audi Autonomous Driving Dataset},''
  2020. [Online]. Available: \url{https://www.a2d2.audi}
\BIBentrySTDinterwordspacing

\bibitem{qi2018}
X.~Qi, Q.~Chen, J.~Jia, and V.~Koltun, ``Semi-parametric image synthesis,''
  \emph{2018 IEEE/CVF Conference on Computer Vision and Pattern Recognition},
  pp. 8808--8816, 2018.

\bibitem{park2019}
T.~Park, M.-Y. Liu, T.-C. Wang, and J.-Y. Zhu, ``Semantic image synthesis with
  spatially-adaptive normalization,'' in \emph{Proceedings of the IEEE
  Conference on Computer Vision and Pattern Recognition}, 2019.

\bibitem{martino2016}
J.~Martino, G.~Facciolo, and E.~Meinhardt-Llopis, ``Poisson image editing,''
  \emph{Image Processing On Line}, vol.~5, pp. 300--325, 11 2016.

\bibitem{tsai2017}
Y.-H. Tsai, X.~Shen, Z.~L. Lin, K.~Sunkavalli, X.~Lu, and M.-H. Yang, ``Deep
  image harmonization,'' \emph{2017 IEEE Conference on Computer Vision and
  Pattern Recognition (CVPR)}, pp. 2799--2807, 2017.

\bibitem{goodfellow2020}
\BIBentryALTinterwordspacing
I.~Goodfellow, J.~Pouget-Abadie, M.~Mirza, B.~Xu, D.~Warde-Farley, S.~Ozair,
  A.~Courville, and Y.~Bengio, ``Generative adversarial networks,''
  \emph{Commun. ACM}, vol.~63, no.~11, p. 139–144, Oct. 2020. [Online].
  Available: \url{https://doi.org/10.1145/3422622}
\BIBentrySTDinterwordspacing

\bibitem{isola2017}
P.~Isola, J.-Y. Zhu, T.~Zhou, and A.~A. Efros, ``Image-to-image translation
  with conditional adversarial networks,'' \emph{2017 IEEE Conference on
  Computer Vision and Pattern Recognition (CVPR)}, pp. 5967--5976, 2017.

\bibitem{zhu2017}
J.-Y. Zhu, T.~Park, P.~Isola, and A.~A. Efros, ``Unpaired image-to-image
  translation using cycle-consistent adversarial networks,'' in \emph{2017 IEEE
  International Conference on Computer Vision (ICCV)}, 2017, pp. 2242--2251.

\bibitem{wang2018pix2pixhd}
T.~Wang, M.-Y. Liu, J.-Y. Zhu, A.~Tao, J.~Kautz, and B.~Catanzaro,
  ``High-resolution image synthesis and semantic manipulation with conditional
  gans,'' \emph{2018 IEEE/CVF Conference on Computer Vision and Pattern
  Recognition}, pp. 8798--8807, 2018.

\bibitem{schonfeld2021}
\BIBentryALTinterwordspacing
E.~Sch{\"o}nfeld, V.~Sushko, D.~Zhang, J.~Gall, B.~Schiele, and A.~Khoreva,
  ``You only need adversarial supervision for semantic image synthesis,'' in
  \emph{International Conference on Learning Representations}, 2021. [Online].
  Available: \url{https://openreview.net/forum?id=yvQKLaqNE6M}
\BIBentrySTDinterwordspacing

\bibitem{gta}
S.~R. Richter, V.~Vineet, S.~Roth, and V.~Koltun, ``Playing for data: {G}round
  truth from computer games,'' in \emph{European Conference on Computer Vision
  (ECCV)}, ser. LNCS, B.~Leibe, J.~Matas, N.~Sebe, and M.~Welling, Eds., vol.
  9906.\hskip 1em plus 0.5em minus 0.4em\relax Springer International
  Publishing, 2016, pp. 102--118.

\bibitem{carla}
A.~Dosovitskiy, G.~Ros, F.~Codevilla, A.~Lopez, and V.~Koltun, ``{CARLA}: {An}
  open urban driving simulator,'' in \emph{Proceedings of the 1st Annual
  Conference on Robot Learning}, 2017, pp. 1--16.

\bibitem{airsim}
\BIBentryALTinterwordspacing
S.~Shah, D.~Dey, C.~Lovett, and A.~Kapoor, ``Airsim: High-fidelity visual and
  physical simulation for autonomous vehicles,'' in \emph{Field and Service
  Robotics}, 2017. [Online]. Available: \url{https://arxiv.org/abs/1705.05065}
\BIBentrySTDinterwordspacing

\bibitem{synthia}
G.~Ros, L.~Sellart, J.~Materzynska, D.~Vazquez, and A.~M. Lopez, ``The synthia
  dataset: A large collection of synthetic images for semantic segmentation of
  urban scenes,'' in \emph{The IEEE Conference on Computer Vision and Pattern
  Recognition (CVPR)}, June 2016.

\bibitem{lee2018}
D.~Lee, S.~Liu, J.~Gu, M.-Y. Liu, M.-H. Yang, and J.~Kautz, ``Context-aware
  synthesis and placement of object instances,'' in \emph{NeurIPS}, 2018.

\bibitem{chen2021}
Y.~Chen, F.~Rong, S.~Duggal, S.~Wang, X.~Yan, S.~Manivasagam, S.~Xue, E.~Yumer,
  and R.~Urtasun, ``Geosim: Realistic video simulation via geometry-aware
  composition for self-driving,'' in \emph{CVPR}, 2021.

\bibitem{nazeri2019}
K.~Nazeri, E.~Ng, T.~Joseph, F.~Qureshi, and M.~Ebrahimi, ``Edgeconnect:
  Generative image inpainting with adversarial edge learning,'' 2019.

\bibitem{Liao2020}
M.~Liao, F.~Lu, D.~Zhou, S.~Zhang, W.~Li, and R.~Yang, ``Dvi: Depth guided
  video inpainting for autonomous driving,'' in \emph{ECCV}, 2020.

\bibitem{yu2019}
J.~Yu, Z.~L. Lin, J.~Yang, X.~Shen, X.~Lu, and T.~S. Huang, ``Free-form image
  inpainting with gated convolution,'' \emph{2019 IEEE/CVF International
  Conference on Computer Vision (ICCV)}, pp. 4470--4479, 2019.

\bibitem{Zhao2020}
C.~Zhao, Q.~Sun, C.~Zhang, Y.~Tang, and F.~Qian, ``Monocular depth estimation
  based on deep learning: An overview,'' \emph{Science China Technological
  Sciences}, pp. 1--16, 2020.

\bibitem{Godard2017}
C.~Godard, O.~{Mac Aodha}, and G.~J. Brostow, ``Unsupervised monocular depth
  estimation with left-right consistency,'' in \emph{CVPR}, 2017.

\bibitem{Alhashim2018}
\BIBentryALTinterwordspacing
I.~Alhashim and P.~Wonka, ``High quality monocular depth estimation via
  transfer learning,'' \emph{arXiv e-prints}, vol. abs/1812.11941, 2018.
  [Online]. Available: \url{https://arxiv.org/abs/1812.11941}
\BIBentrySTDinterwordspacing

\bibitem{Pillai2019}
S.~Pillai, R.~Ambrus, and A.~Gaidon, ``Superdepth: Self-supervised,
  super-resolved monocular depth estimation,'' 05 2019, pp. 9250--9256.

\bibitem{Lee2019}
J.~H. Lee, M.-K. Han, D.~W. Ko, and I.~H. Suh, ``From big to small: Multi-scale
  local planar guidance for monocular depth estimation,'' \emph{arXiv preprint
  arXiv:1907.10326}, 2019.

\bibitem{hu2020penet}
M.~Hu, S.~Wang, B.~Li, S.~Ning, L.~Fan, and X.~Gong, ``Penet: Towards precise
  and efficient image guided depth completion,'' 2021.

\bibitem{connected_components}
\BIBentryALTinterwordspacing
C.~Fiorio and J.~Gustedt, ``Two linear time union-find strategies for image
  processing,'' \emph{Theoretical Computer Science}, vol. 154, no.~2, pp.
  165--181, 1996. [Online]. Available:
  \url{https://www.sciencedirect.com/science/article/pii/0304397594002622}
\BIBentrySTDinterwordspacing

\bibitem{hu_moments}
M.-K. Hu, ``Visual pattern recognition by moment invariants,'' \emph{IRE
  transactions on information theory}, vol.~8, no.~2, pp. 179--187, 1962.

\bibitem{adamoptim}
\BIBentryALTinterwordspacing
D.~P. Kingma and J.~Ba, ``Adam: {A} method for stochastic optimization,'' in
  \emph{3rd International Conference on Learning Representations, {ICLR} 2015,
  San Diego, CA, USA, May 7-9, 2015, Conference Track Proceedings}, Y.~Bengio
  and Y.~LeCun, Eds., 2015. [Online]. Available:
  \url{http://arxiv.org/abs/1412.6980}
\BIBentrySTDinterwordspacing

\bibitem{FID}
M.~Heusel, H.~Ramsauer, T.~Unterthiner, B.~Nessler, and S.~Hochreiter, ``Gans
  trained by a two time-scale update rule converge to a local nash
  equilibrium,'' in \emph{Proceedings of the 31st International Conference on
  Neural Information Processing Systems}, ser. NIPS'17.\hskip 1em plus 0.5em
  minus 0.4em\relax Red Hook, NY, USA: Curran Associates Inc., 2017, p.
  6629–6640.

\bibitem{drn}
F.~Yu, V.~Koltun, and T.~Funkhouser, ``Dilated residual networks,'' in
  \emph{Computer Vision and Pattern Recognition (CVPR)}, 2017.

\bibitem{kitti_depth}
J.~Uhrig, N.~Schneider, L.~Schneider, U.~Franke, T.~Brox, and A.~Geiger,
  ``Sparsity invariant cnns,'' in \emph{International Conference on 3D Vision
  (3DV)}, 2017.

\end{thebibliography}

\end{document}